\batchmode
\makeatletter
\def\input@path{{\string"/Users/andrewpowell/iCloud Drive (Archive)/Documents/\string"}}
\makeatother
\documentclass[oneside,english]{amsart}
\usepackage[T1]{fontenc}
\usepackage[latin9]{inputenc}
\usepackage{amsthm}

\makeatletter
\numberwithin{equation}{section}
\numberwithin{figure}{section}

\makeatother

\usepackage{babel}
\begin{document}
\title{Artificial Consciousness and Security}
\author{Andrew Powell}
\address{Dr. Andrew Powell, Honorary Senior Research Fellow, Institute for
Security Science and Technology, Level 2 Admin Office Central Library,
Imperial College London, South Kensington Campus, London SW7 2AZ,
United Kingdom.}
\email{andrew.powell@imperial.ac.uk}
\begin{abstract}
This paper describes a possible way to improve computer security by
implementing a program which implements the following three features
related to a weak notion of artificial consciousness: (partial) self-monitoring,
ability to compute the truth of quantifier-free propositions and the
ability to communicate with the user. The integrity of the program
could be enhanced by using a trusted computing approach, that is to
say a hardware module that is at the root of a chain of trust. This
paper outlines a possible approach but does not refer to an implementation
(which would need further work), but the author believes that an implementation
using current processors, a debugger, a monitoring program and a trusted
processing module is currently possible.
\end{abstract}

\maketitle

\section{Introduction}

It is plausible to believe that the minimum condition that distinguishes
an artificially conscious computer from one that is not conscious
is the ability to self-monitor (an idea that is evident in the postscript
to \cite{key-1}, and is explored in, for example, \cite{key-1-1}).
There are many other approaches to artificial consciousness, from
flat out denial of its possibility (usually based on a belief in a
fundamentally qualitative type of existence, qualia, which inanimate
things do not possess), through the views that consciousness is only
to be associated with biological systems, that consciousness must
be associated with language, that a conscious being must have an internal
representation of itself, that consciousness is an emergent property
of sufficiently complex systems, to the view above that consciousness
is a function of the ability to self-refer or self-monitor and the
even stronger view that consciousness is a property of computations.
\cite{key-3} contains a range of reasonably current views on artificial
consciousness and \cite{key-3-0-1} and \cite{key-3-0-2} are recent
surveys. This paper does not provide a critique of various views of
artificial consciousness, but instead advocates that three criteria
should be tested empirically to investigate their adequacy for artificial
consciousness in order to establish a possibly weak notion of artificial
consciousness which can be used to improve the security of computer
systems. \\
\\
The three criteria are:
\begin{itemize}
\item Self-monitoring
\item Ability to make judgements
\item Ability to communicate, \emph{i.e.} at minimum respond with ``yes''
and ``no'' answers to external questions
\end{itemize}
To be clear, these criteria are not seen as an anything more than
framing an hypothesis, which will need help from techniques in machine
learning, expert systems and machine visual representation in order
to be tested. It may be that an autonomous agent approach with consciousness
as the broker between the activities of the agents (see for example
the LIDA parallel processing model of S. Franklin, for example \cite{key-3-1,key-3-1-1,key-3-1-2},
or the neural network agents model of M. Shanahan, see \cite{key-3-1-3,key-3-1-4},
or the neural network state machine approach of I. Aleksander (see
\cite{key-3-2}) may produce machines which pass the test of consciousness
as expressed by the three criteria above. There is merit in the view
that an artificially conscious entity does need representations of
objects in the world around us to make judgments that are decidable
by others (see Section 4), but in this paper it is argued that judgements
about the state\footnote{In this paper, \emph{state} will refer to the values of all registers
and the current instruction executed at a specific time. } of an entity's own registers and memory has value for the purpose
of maintaining security. \\
\\
In this paper the focus will be on a computer as the artificially
conscious entity and on the security implications of artificial consciousness,
which exist even for a weak notion of self-monitoring. Section 2 discusses
how far self-monitoring can be achieved. Section 4 explains why the
ability to make judgements and the ability to communicate are reasonable
criteria for artificial consciousness and how judgements and communications
should be characterized. Sections 3 and 5 discuss the implications
for security of self-monitoring, judgment and communications of an
artificially conscious program and connections to the trusted computed
initiative implemented in recent computing platforms. In summary then,
the notions of communications, self-monitoring and judgement for a
computer are explored and their implications for improved security
of computer systems are considered. 

\section{Self-monitoring}

\emph{Self-monitoring} is not the same as self-representation but
does imply some kind of self-awareness. \emph{Self-representation}
as a notion implies that a computer program has a model of the self
which it uses to validate references in judgements about actions made
by the self, but self-monitoring on the other hand only requires that
the program can check (some of) its own activities. Although self-representation
is not explicitly pursued in this paper, it may have utility as an
aspect of a consciousness because it is reasonable to suppose that
the self can also be represented to the self. However, self-monitoring
is taken to be the more fundamental notion, as there are limits to
how faithful a representation of self can be (see the comments below
about the limits of self-monitoring). To be more precise about self-monitoring,
a computer program \emph{self-monitors} if it is capable of checking
the values of all the registers or variables that it uses and the
instruction that the program is currently executing and the history
of the state of the program (that is, the register values and instructions
executed, indexed by time). A computer process \emph{self-monitors}
if it self-monitors as a computer program and it is able to monitor
the state of any interrupt sent to it by another process.\\
\\
It is possible for a program or a process to self-monitor (if it has
not crashed)\footnote{A process has crashed if the program has not terminated and its execution
cycles through a sequence of states. } because the self-monitoring function is very similar to what a debugger
or instrumentation program does. Strictly, a debugger is a program
which enables any other program to be monitored without changing its
computation steps or the values of its variables; and it is of course
true there will be registers and values whose current values cannot
be read, \emph{i.e.} those registers which are used to check the values
of registers of the instrumented program. Recent approaches to instrumenting
a program by dynamically patching its execution path are given at
\cite{key-5,key-6,key-4-1}. In the present context such a limitation
is acceptable for two reasons. Firstly, self-monitoring is useful
insofar as it concerns monitoring the status of processes spawned
by a given process, for example, to check if the processes have crashed
or are stuck in an endless loop, rather than trying to determine whether
the process itself is stuck in an endless loop (which is in general
is unsolvable because it is equivalent to the halting problem\footnote{This well known result is due to A. Turing \cite{key-4}, but a modern
approach is to define $f(e)=1$ if $(\exists x\in N)(\{e\}(e)=x)$
and $f(e)=0$ otherwise,\emph{ i.e.} if $\{e\}(e)$ is not defined,
where $e$ is the natural number code of (the syntax of) a program
and $\{e\}:N\rightarrow N$ is the function that the program implements,
assumed to be a natural number function. Then if $f$ were computable,
then $f=\{h\}$ for some numerical program code $h$, and it follows
that $\{h\}(h)=0$ if $\{h\}(h)$ is not defined, contradiction.}). Secondly, it is possible to check the value of all registers and
the currently executing instruction with a time delay, as previously
executed instructions and register values can be archived. It is also
possible for the values of the internal state of a process to be monitored
by a separate process which can set a flag in one registers used by
the first process if the first process is behaving abnormally, with
the limitation that the first process cannot reciprocally monitor
the monitoring processing (which would be equivalent to self-monitoring).
So, for practical purposes programs and processes can self-monitor.
More theoretically, it is unreasonable to expect a monitoring interface
to expose everything about the monitoring program. Humans rely on
indirect reports from sensors, whereas an artificially conscious computer
could expose far more of its hardware state as well as the states
of its processes.\\
\\
The implications of a self-monitoring program are that the operating
system (which is a management program after all) can monitor all the
programs and processes that the operating system manages and can intervene
if they crash, do not respond to interrupts, or just behave abnormally
(take up a lot of memory or get stuck in a loop with no change in
the value of the variables in the loop condition). 

\section{Security implications of self-monitoring}

If it is possible to tell whether a process is behaving abnormally,
it is plausible to believe that the operating system can check whether
a process or program other than itself is operating insecurely. Of
course security is difficult to define in general because security
is relative to a set of specific security policies\footnote{There is a view in  \cite{key-6-3} that security properties of computer
programs are properties of sets of execution paths (or traces), constraining
those sets of traces in some way and specifying which systems (sets
of traces) a security policy relates to. This is an elegant way of
formulating security properties and formalizing security policies.}, but in terms of vulnerabilities not envisaged by the programmer,
security means that the variables have the values expected and that
no memory structures used in the management of program execution are
modified other than by the operating system. A value of a variable
may be said to be \emph{expected} if the program assigns data types
to variables and the value is in the range associated with the data
type. In the case where the operating system instruments every program
by assigning data types to all variables, checks whether all values
of the variables are expected, and manages all access to memory structures,
it can be seen that a self-monitoring operating system could identify
programs which are operating insecurely and could instrument them
in such a way that evidence could be provided to a system administrator
so that the insecure program could be closed down.\\
\\
In fact, we can even be bolder in our claims about what a self-monitoring
operating system could monitor. It could monitor attempts to modify
operating system functions and libraries but, in general, not attempted
changes to instrumentation of those functions. It would be only be
possible to address the risk of unauthorized modification of the operating
system if there was a hierarchy of trust. If the operating system
is trustworthy then it could assign a trustworthiness rating to programs
based on the number of security reports raised. It would be particularly
useful to combine this approach with a trusted computing approach\footnote{See for example \cite{key-6-1}. Trusted computing has been implemented
on the motherboards of some business-focussed personal computers and
mobile computing platforms.} which uses hardware separation (\emph{i.e.} trusted processing modules)
to verify the integrity of the operating system, to provide a chain
of trust of programs run on the system, and to manage (via a virtualization
layer) the execution of any programs (whether standard user programs
or high privilege programs such as kernel loadable modules). The trusted
computing module would prevent programs from having an impact if they
execute insecure code and the security verification will check the
trustworthiness of the program.

\section{Judgement and communication as criteria of artificial consciousness}

An artificial consciousness that could make judgements for itself
would reduce the decision-making burden on the user. The practical
reason for including judgements in the criteria for consciousness
is that it seems impossible to make decisions about what you are monitoring
without the ability to make judgements. Thanks to a line of logicians
from G. Frege (see \cite{key-8})\footnote{Arguably the line originates from I. Kant (see \cite{key-2}) and
includes E. Husserl.} onwards, we understand what it is to make a judgement about a set
of concepts and objects. That is, we can in principle decide (\emph{i.e.}
compute the truth or falsehood of) a statement that does not contain
unbounded logical quantifiers (such as ``for all'' or ``there exists'')\footnote{In general dedidable propositions with quantfication over an euumerable
set $S$ are of the form $(\exists y)P(y,x)$ if $x\in S$ and $(\exists y)Q(y,x)$
if $x\notin S$, but most such propositions will not be decidable
by a computer with fixed finite resources. } but may contain logical operators such as ''and'', ``or'', ``not''
and ``implies''. A judgment is then a computable function (that
is, a computation) from properties (or predicates) and objects into
the set that contains ``true'' and ``false''. To be clear, we
can apply a program to a (natural number) code of a property that
could apply to a set of (codes of) input objects, and compute whether
the property applies to a given set of objects or not. For example,
if we wish to decide where natural number $c$ satisfies the natural
number relation $a\le x\le b$ we could code $a\le x\le b$ as $\left\lceil a\le x\le b\right\rceil $
using a computable coding $\left\lceil \right\rceil $ and then substitute
$c$ for $x$ in $\left\lceil a\le x\le b\right\rceil $ and decide
the truth of $\left\lceil a\le c\le b\right\rceil $ by means of a
computable function. Of course, the types of judgement that a computer
can make will concern objective states of affairs that it can represent,
such as whether a process is or is not responding to interrupts (in
a certain timescale), but not be about wishes or intentions.\\
\\
It is worth stressing that the operating systems should make judgements
about all programs that the operating system manages (at least in
the form of recommendations to an administrator) on a frequent basis
in terms of program health (where they are caught in an endless loop
or require too much system resources to run), program security (against
security policies and needing to pass vulnerability checks) and program
safety (against safety policies and needing to pass vulnerability
checks). The idea is that the operating system would run through all
the tasks it needs to perforn and compute each as a judgment, recording
the results of the judgements in a log.\\
\\
It is also worth noting that not all properties are computably decidable\footnote{In general arithmetical predicates containing any unbounded natural
number quantifiers are not decidable by means of a computation unless
they have the specific form noted in Footnote 7. } and some are not practically computably decidable (because any computation
has a long run-time), but it is nevertheless possible to decide whether
for example a set of bits (a pixel) represents the colour blue, whether
a certain shape could represent a cat, or whether indeed a certain
process has not responded to an interrupt.\footnote{These examples are deliberately taken mainly from machine learning
of visual representations because that area provides a rich source
of decidable judgements. Speech analytics is another such area, as
is of course the content of the computer's own registers.} If we allow deep learning neural networks\footnote{Layers in convolutional neural networks, where the convolution operation
picks out features, naturally form a hierarchy of increasing abstraction.}, it is possible to represent concepts of differing levels of abstraction
and to classify objects under those concepts. In order to decide properties
and make judgements, the artificially conscious program will need
to write and run programs of its own,\emph{ i.e.} spawn processes.
In order to make judgements about propositions that are not decidable
but which are theorems of axiom systems, we might want to allow the
artificial consciousness to deduce theorems from (codes of) axioms
using inference rules, understand the axioms by verifying that the
axioms have a model\footnote{Ideally we would want to show that the axioms are true in a particular
model of the axioms, because truth in some model shows the consistency
of the axioms. In order to build models computable representations
of arbitrary elements and defined functions/predicates of the model
will be needed. \cite{key-9} is a very readable account of model
theory, using games with finite rule sets to build models.}, and even to be able to propose new axioms by producing models which
satisfy those axioms (perhaps by using neural networks to classify
propositions as ``theorems'' or not). However, even without making
the artificially conscious program into a logician or a data scientist,
the value of being able to make judgments is considerable in terms
of the ability of the artificial consciousness to enforce security
and safety policies and to improve clarity and efficiency of interacting
with the operating system for the user.\\
\\
The reason why a program that can make judgements results in greater
clarity for the user is that the computations of decidable propositions
will form a justification of the decisions that the program recommends.
This approach will also increase efficiency for the user because the
operating system can make recommendations to the user or take actions
in a way that does not cause the user to try to guess at the cause
of messages from the operating system.\\
\\
The ability to communicate is included as a criterion of artificial
consciousness as a minimum condition for testing artificial consciousness,
otherwise an external user will have to monitor its state directly.
An artificially conscious program needs to communicate with the programs
that it monitors, and operating systems are expected to report to
users on the the status of programs that are running and to implement
the users' commands. The ability to communicate interactively and
faithfully would be desirable. For these reasons, the ability to communicate
to other programs and to users is essential. At minimum that communication
could be ``yes'' or ``no'' \emph{(i.e.} one bit of information),
although in practice data and functions of all types could be passed
through the program's interface, including validation of the accuracy
of the information communicated.

\section{Security implications of judgement and communications}

The ability to form judgements could be used to prevent a user making
mistakes and in coming to evidence-based decisions. When combined
with the trusted computing techniques noted in connection with self-monitoring,
the artificially conscious program would have some evidence for the
integrity of its own functioning and for the soundness of its own
judgements. The ability to communicate on the other hand could introduce
vulnerabilities into the program if the types and validity of the
value of program inputs are not checked, and the integrity of the
messages communicated would need to be assured (by cryptographic means
for example). Communication is necessary for the worth of the artificially
conscious program to be realized in terms of helping a user make decisions
and to report back information. In any case vulnerabilities in programs
through specially crafted inputs are not new for any operating system,
nor is the need for integrity checking of communications. Trusted
computing could also provide integrity checking of communications.

\section{Conclusions}

In this paper an approach to artificial consciousness is suggested
which is sufficient for increasing the security of operating systems,
namely communications, judgments and self-monitoring. It is also suggested
that self-monitoring brings significant security benefits in supporting
the termination of programs which do not respond to interrupts or
otherwise exhibit unusual behaviour, that communications is necessary
for testing the functioning of an artificially conscious program,
and that the ability to make judgements is useful for user decision-support.

\end{document}